\documentclass[conference]{IEEEtran}

\makeatletter
\def\ps@IEEEtitlepagestyle{%
  \def\@oddfoot{\mycopyrightnotice}%
  \def\@evenfoot{}%
}
\def\mycopyrightnotice{%
  {\footnotesize XXX-X-XXXX-XXXX-X/XX/\$XX.00~\copyright~20XX IEEE\hfill}
  \gdef\mycopyrightnotice{}
}

\usepackage{blindtext}
\usepackage{eso-pic}
\IEEEoverridecommandlockouts
\usepackage{tabularx}
\usepackage{balance}
\usepackage{lipsum}
\usepackage{hhline}
\usepackage{multirow}
\usepackage{makecell}
 \usepackage{booktabs}
\usepackage{array, tabularx}
\usepackage{cite}
\usepackage{amsmath,amssymb,amsfonts}
\usepackage{algorithmic}
\usepackage{graphicx}
\usepackage{textcomp}
\usepackage{xcolor}
\def\BibTeX{{\rm B\kern-.05em{\sc i\kern-.025em b}\kern-.08em
    T\kern-.1667em\lower.7ex\hbox{E}\kern-.125emX}}
    
\usepackage{eso-pic}
\newcommand\AtPageUpperMyright[1]{\AtPageUpperLeft{%
 \put(\LenToUnit{0.17\paperwidth},\LenToUnit{-2cm}){%
     \parbox{0.9\textwidth}{\raggedleft\fontsize{8}{11}\selectfont #1}}%
 }}%
\newcommand{\conf}[1]{%
\AddToShipoutPictureBG*{%
\AtPageUpperMyright{#1}
}
}

\begin{document}
\title{\vspace*{1cm} TinyML for Acoustic Anomaly Detection in IoT Sensor Networks\\

\thanks{}
}

\author{\IEEEauthorblockN{Amar Almaini}
\IEEEauthorblockA{\textit{computer science} \\
\textit{Deggendorf Institute of Technology}\\
Deggendorf, Germany \\
amar.almaini@th-deg.de}
\and
\IEEEauthorblockN{Jakob Folz}
\IEEEauthorblockA{
\textit{Computer Science and Engineering}\\
\textit{University of West Bohemia} \\
Pilsen, Czech Republic\\
jfolz@kiv.zcu.cz
}
\and
\IEEEauthorblockN{Ghadeer Ashour}
\IEEEauthorblockA{\textit{The Applied College} \\
\textit{King Abdulaziz University}\\
Jeddah, Saudi Arabia \\
Gaashor@kau.edu.sa}

}

\maketitle
\conf{\textit{  Proc. of the International Conference on Electrical, Computer, Communications and Mechatronics Engineering (ICECCME 2025) \\ 
16-19 October 2025, Zanzibar, Tanzania}}

\begin{abstract}
Tiny Machine Learning enables real-time, energy-efficient data processing directly on microcontrollers, making it ideal for Internet of Things sensor networks. This paper presents a compact TinyML pipeline for detecting anomalies in environmental sound within IoT sensor networks. Acoustic monitoring in IoT systems can enhance safety and context awareness, yet cloud-based processing introduces challenges related to latency, power usage, and privacy. Our pipeline addresses these issues by extracting Mel Frequency Cepstral Coefficients from sound signals and training a lightweight neural network classifier optimized for deployment on edge devices. The model was trained and evaluated using the UrbanSound8K dataset, achieving a test accuracy of 91\% and balanced F1-scores of 0.91 across both normal and anomalous sound classes. These results demonstrate the feasibility and reliability of embedded acoustic anomaly detection for scalable and responsive IoT deployments.
\end{abstract}


\section{Introduction}
The widespread adoption of Internet of Things (IoT) devices has enabled scalable sensor networks across domains such as environmental monitoring, infrastructure management, and smart city applications. Among the available sensing modalities, acoustic sensing offers valuable insights for detecting anomalies such as mechanical malfunctions, intrusions, or safety hazards. However, processing audio data in traditional cloud-based systems often incurs latency, requires high communication bandwidth, and raises privacy concerns. These limitations are especially relevant in distributed or resource-constrained IoT deployments, where constant data transmission is not feasible. Tiny Machine Learning (TinyML) has emerged as a promising solution to this challenge. By enabling machine learning inference directly on microcontrollers, TinyML allows sensor nodes to process data locally with low latency and power consumption, while preserving privacy by keeping raw data on the device \cite{alajlan2024tinyml}. This approach is well suited for acoustic anomaly detection, where timely and context-aware decisions are critical. Recent studies have explored the potential of TinyML in this domain. Hammad et al.\ proposed an unsupervised LSTM autoencoder for detecting noise anomalies and successfully deployed it on an ESP32 microcontroller \cite{hammad2023unsupervised}. Similarly, Botero-Valencia et al.\ developed a low-cost environmental monitoring station using TinyML to analyze sound, light, and air quality data at the edge \cite{botero2023lowcost}. Räni \cite{rani2022noiseclassification} evaluated the classification of urban noise events using MFCC features and embedded classifiers, offering insight into real-time inference on microcontrollers. These works demonstrate the viability of deploying compact models on low-power hardware but often rely on specific sensor platforms or custom hardware setups. In this paper, we present a TinyML pipeline for binary sound anomaly detection in sensor networks, trained and evaluated on the publicly available UrbanSound8K dataset. The system extracts Mel Frequency Cepstral Coefficients (MFCCs) from urban audio recordings and uses a compact neural network classifier designed for embedded deployment. The model is quantized and converted to TensorFlow Lite Micro format to ensure compatibility with microcontroller platforms. Our approach achieves a test accuracy of 91\% with balanced precision and recall, confirming the feasibility of using lightweight audio classifiers for real-time, edge-based anomaly detection. By focusing on a reproducible and generalizable setup, this work offers a practical and scalable foundation for integrating acoustic intelligence into resource-constrained IoT environments. In this work, we frame anomaly detection as a binary sound classification problem, where "anomalous" events refer to disruptive or urgent urban sounds (e.g., sirens, gunshots) that deviate from typical ambient noise. While traditional anomaly detection often identifies outliers without labeled classes, our approach explicitly uses labeled categories to enable targeted detection of critical events in urban IoT contexts. The remainder of this paper is structured as follows: Section~\ref{Related Work} reviews related work on TinyML and sound-based anomaly detection. Section~\ref{System Architecture} describes the system architecture. Section~\ref{Evaluation} presents the evaluation results. Section~\ref{Conclusion}~concludes~and~outlines~ future~work.

\section{Related Work} \label{Related Work}
TinyML has emerged as a transformative approach to enabling on-device intelligence in IoT sensor networks. By allowing microcontrollers to process sensor data locally, TinyML facilitates low-latency response, reduced energy consumption, and improved privacy. Tsoukas et al.~\cite{tsoukas2024review} provide a comprehensive review of the TinyML landscape, highlighting trends such as model compression, quantization, and edge deployment for real-world applications. Elhanashi et al.~\cite{elhanashi2024advancements} discuss the limitations and capabilities of TinyML in practical IoT scenarios, emphasizing its growing relevance for monitoring, classification, and anomaly detection in resource-constrained environments. Sound-based anomaly detection is one of the most compelling applications of TinyML in IoT. Hammad et al.~\cite{hammad2023unsupervised} proposed an unsupervised approach based on LSTM autoencoders for detecting environmental noise anomalies. Their solution, implemented on an ESP32-S3 microcontroller, demonstrates the feasibility of on-device anomaly detection using reconstruction error. However, their system requires real-world sensor data and deployment infrastructure, which may limit its replicability in early experimental stages. Complementing this, Botero-Valencia et al.~\cite{botero2023lowcost} developed a low-cost environmental monitoring station capable of measuring air quality, sound levels, and light levels. Their prototype leverages TinyML for edge inference, illustrating the benefits of decentralized intelligence in IoT systems. Räni~\cite{rani2022noiseclassification} explored the classification of urban noise sources such as traffic, industrial, and human sounds using MFCC features and small neural networks. The work compares Google's Micro pre-processing and MFCCs on an EFR32 microcontroller platform, offering insights into embedded audio classification and deployment challenges. However, reported accuracy levels were modest, and the evaluation was conducted on limited sensor setups. From a model architecture perspective, Cerutti et al.~\cite{Cerutti2020compactrnn} proposed compact recurrent neural networks for acoustic event detection, optimized for execution on embedded platforms. Their work demonstrates that accurate temporal modeling is feasible even with strict memory and power constraints. Muthumala et al.~\cite{muthumala2024comparison} conducted a comparative study of TinyML techniques for analyzing embedded acoustic emission signals. Their findings reveal the trade-offs between inference performance and computational efficiency, relevant to scenarios involving anomaly detection in sensor networks. Recent contributions such as the DCASE 2024 Challenge Task 2 have explored unsupervised acoustic anomaly detection in industrial settings, providing valuable benchmarks and encouraging advancements in anomaly-aware sound monitoring~\cite{nishida2024description}.  On a broader level, DeMedeiros et al.~\cite{demedeiros2023survey} provide a survey of AI-based anomaly detection approaches in IoT and wireless sensor networks. Their review classifies methods by application domain, model type, and system architecture, reinforcing the growing significance of edge-based intelligence for anomaly-aware systems. A complementary survey by Chatterjee and Ahmed~\cite{chatterjee2022iot} examines anomaly detection across the IoT landscape, including statistical, ML-based, and hybrid approaches. UrbanSound8K is a widely adopted dataset for urban sound classification,  offering over 8,000 labeled recordings across diverse environmental categories~\cite{salamon2014dataset}. Its standardized folds support reproducibility in model evaluation. Other benchmark datasets such as ESC-50 also provide a standardized collection of labeled environmental sounds and are widely used for model comparison and evaluation~\cite{piczak2015esc}. In terms of feature extraction, MFCCs remain a standard feature representation for embedded sound classification due to their compactness and efficiency on low-power hardware \cite{rani2022noiseclassification}. In addition to MFCC-based systems, resource-adaptive convolutional architectures have also shown promise for fast and efficient environmental sound classification~\cite{fang2022fast}. Compared to prior work, our contribution integrates these established techniques into a fully reproducible TinyML pipeline based on UrbanSound8K. We achieve a final test accuracy of 91\% using a lightweight neural network trained on MFCC features. The model is quantized for deployment on microcontrollers and designed to fit real-world constraints without sacrificing classification performance. This balance of accuracy, portability, and reproducibility represents a meaningful step forward in practical edge-based anomaly detection.

\section{System Architecture and Implementation} \label{System Architecture}
\begin{figure}[h]
  \centering
  \includegraphics[width=\linewidth]{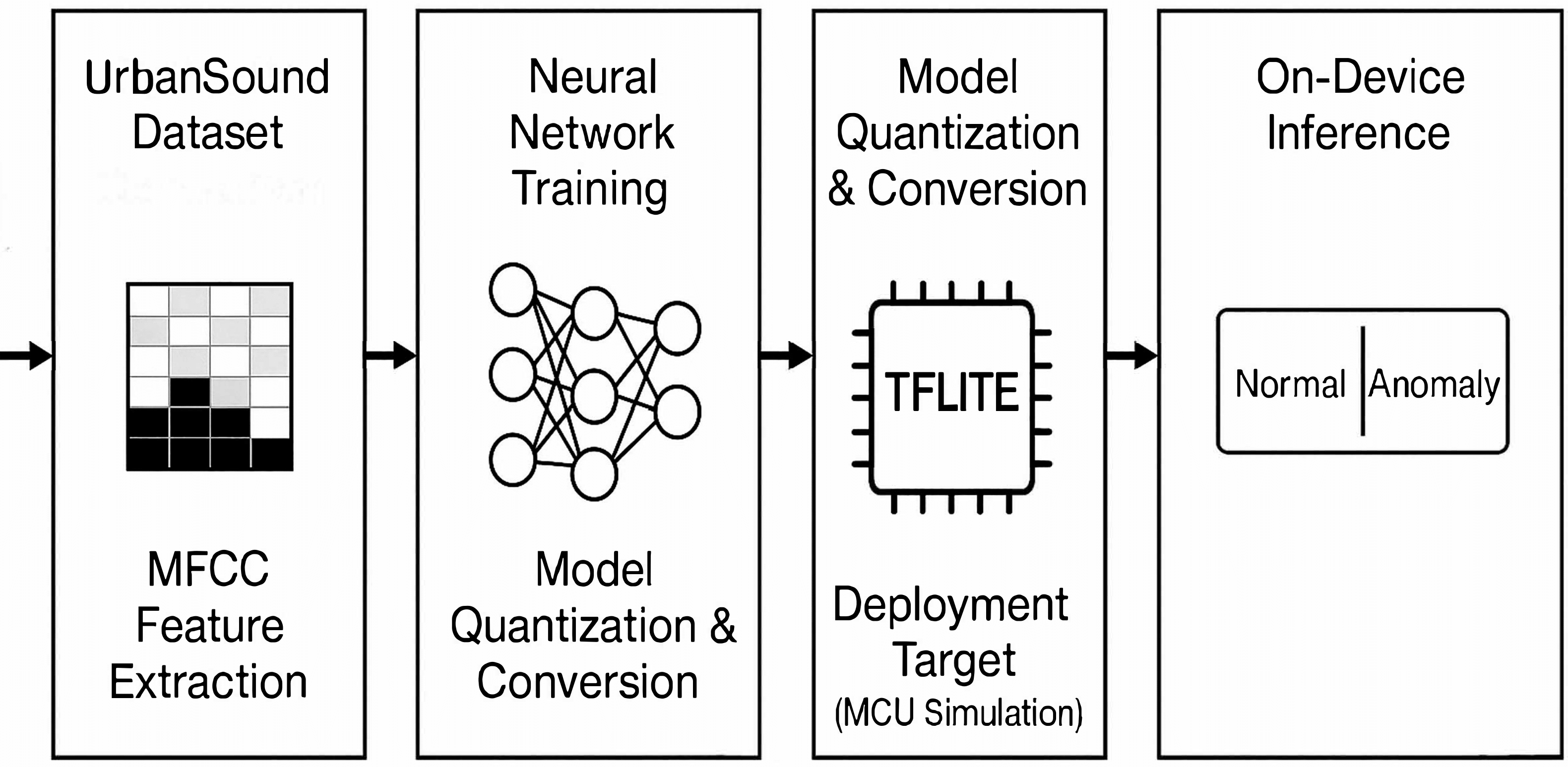}
  \caption{TinyML pipeline for MFCC-based sound anomaly detection on IoT sensor nodes.}
  \label{fig:architecture}
\end{figure}

The proposed system implements an end-to-end TinyML pipeline for acoustic anomaly detection in IoT sensor networks using the publicly available UrbanSound8K dataset. As illustrated in Fig.~\ref{fig:architecture}, the pipeline consists of four primary stages: (1) MFCC-based feature extraction, (2) training of a lightweight neural network classifier, (3) model quantization and optimization, and (4) deployment preparation using TensorFlow Lite for Microcontrollers (TFLM).

\noindent The UrbanSound8K dataset contains 8,732 annotated audio samples spanning 10 urban sound classes, including engine idling, air conditioners, jackhammers, sirens, and gunshots~\cite{salamon2014dataset}. To frame this as a binary classification problem, we grouped the original classes into two categories: \textit{normal} sounds (e.g., engine idling, air conditioner, children playing) and \textit{anomalous} sounds (e.g., siren, gunshot, jackhammer, dog bark). Each audio clip was resampled to 16~kHz and trimmed or padded to a fixed duration. MFCCs were extracted using 40\,ms frames with 50\% overlap, yielding 13 coefficients per frame. This resulted in compact 2D feature maps per clip, which were flattened into 1D vectors for input into a dense model. MFCCs remain a widely used representation for audio classification tasks in embedded settings due to their efficiency and low computational overhead~\cite{rani2022noiseclassification}. The classification model is a fully connected neural network with two hidden layers, each followed by dropout to mitigate overfitting. The final output layer uses a sigmoid activation function to perform binary classification. Training was performed using the Adam optimizer and binary cross-entropy loss, with hyperparameters listed in Table~\ref{tab:hyperparams}. Although configured for 50 epochs, training was halted early after 36 epochs due to validation accuracy plateauing. The best-performing model was subsequently quantized to 8-bit integers and converted into TensorFlow Lite format for compatibility with TFLM. While on-device deployment is beyond the scope of this work, the model's footprint and structure were optimized for embedded microcontrollers. The final quantized model contains approximately 62,000 parameters (exactly 61,825), occupying around 60 kB of memory after 8-bit quantization. Although designed to run on general microcontroller platforms (e.g., ARM Cortex-M series), exact resource utilization and power consumption measurements will be analyzed in future work. The pipeline is fully reproducible, hardware-agnostic, and generalizable for anomaly detection in resource-constrained IoT settings.\\
\noindent
\textbf{Implementation Details:} The system was implemented in Python 3.10 using TensorFlow 2.15 and \texttt{librosa} 0.10.1 for audio preprocessing. Training and evaluation were conducted on a standard workstation (Intel Core i7, 16\,GB RAM) without GPU acceleration. Our pipeline design was further guided by practical deployment examples such as Mistry’s Raspberry Pi RP2040-based TinyML audio classification tutorial~\cite{mistry2021tinyml}.

\begin{table}[h]
\setlength{\tabcolsep}{20pt} 
\centering
\caption{Model training and preprocessing parameters.}
\label{tab:hyperparams}
\begin{tabularx}{\linewidth}{ll}
\toprule
\textbf{Parameter} & \textbf{Value} \\
\midrule
MFCC coefficients & 13 \\
Frame length & 40 ms \\
Frame overlap & 50\% \\
Sample rate & 16 kHz \\
Input shape & 13 $\times$ 32 \\
Hidden layers & [128, 64] (ReLU) \\
Dropout rate & 0.2 \\
Output layer & Sigmoid \\
Loss function & Binary cross-entropy \\
Optimizer & Adam \\
Initial learning rate & 0.001 \\
Learning rate scheduler & ReduceLROnPlateau \\
Early stopping & Patience = 5, restore best weights \\
Batch size & 32 \\
Max epochs & 50 \\
Actual epochs used & 36 (early stopping) \\
Batch size & 64 \\
Validation split & 20\% \\
\bottomrule
\end{tabularx}
\end{table}

\section{Evaluation} \label{Evaluation}
We evaluated the proposed TinyML-based anomaly detection pipeline on the full UrbanSound8K dataset, consisting of 8,732 labeled audio samples across 10 urban sound classes. These were grouped into two categories: \textit{normal} and \textit{anomalous}, aligning with the practical goal of detecting out-of-place or high-salience acoustic events in sensor network deployments. After preprocessing and MFCC extraction, the data was split using a stratified 80/20 train-test ratio, yielding 6,985 training and 1,747 testing samples. A lightweight fully connected neural network was trained using TensorFlow for up to 50 epochs, with early stopping halting the process after 36 epochs. To assess the impact of quantization, we evaluated both the original floating-point model and the quantized TinyML model using the same criteria: accuracy, F1-score, ROC AUC, and average precision. As shown in Table~\ref{tab:model_comparison}, the original model achieved 95\% accuracy and an F1-score of 0.95, with very high discriminative power (ROC AUC = 0.991, Avg. Precision = 0.992). The quantized model, optimized for deployment on microcontrollers via TensorFlow Lite Micro, achieved 91\% accuracy and an F1-score of 0.91, with only minor drops in AUC and average precision. This trade-off is acceptable in constrained IoT scenarios where energy efficiency and memory footprint are critical.

\begin{table}[h]
\setlength{\tabcolsep}{5pt} 
\centering
\caption{Model Performance Comparison Before and After Quantization}
\label{tab:model_comparison}
\begin{tabularx}{\linewidth}{lcccc}
\toprule
\textbf{Model} & \textbf{Accuracy} & \textbf{F1-Score} & \textbf{ROC AUC} & \textbf{Avg. Precision} \\
\midrule
Original (float32)   & 0.95 & 0.95 & 0.991 & 0.992 \\
Quantized (int8)     & 0.91 & 0.91 & 0.970 & 0.970 \\
\bottomrule
\end{tabularx}
\end{table}

\noindent To further understand the behavior of the quantized model, we analyzed its class-wise prediction characteristics. Table~\ref{tab:metrics} shows that performance was consistent across both classes, with precision and recall values above 0.88.

\begin{table}[htbp]
\setlength{\tabcolsep}{10pt} 
\centering
\caption{Class-wise Evaluation Metrics for the Quantized Model}
\label{tab:metrics}
\begin{tabularx}{\linewidth}{lcccc}
\toprule
\textbf{Class} & \textbf{Precision} & \textbf{Recall} & \textbf{F1-Score} & \textbf{Support} \\
\midrule
Normal (0)     & 0.88               & 0.94            & 0.91              & 3996             \\
Anomalous (1)  & 0.94               & 0.89            & 0.91              & 4283             \\
\midrule
\textbf{Accuracy} & –               & –               & \textbf{0.91}     & 8279             \\
\bottomrule
\end{tabularx}
\end{table}

\begin{figure}[htbp]
    \centering
    \includegraphics[width=0.9\linewidth]{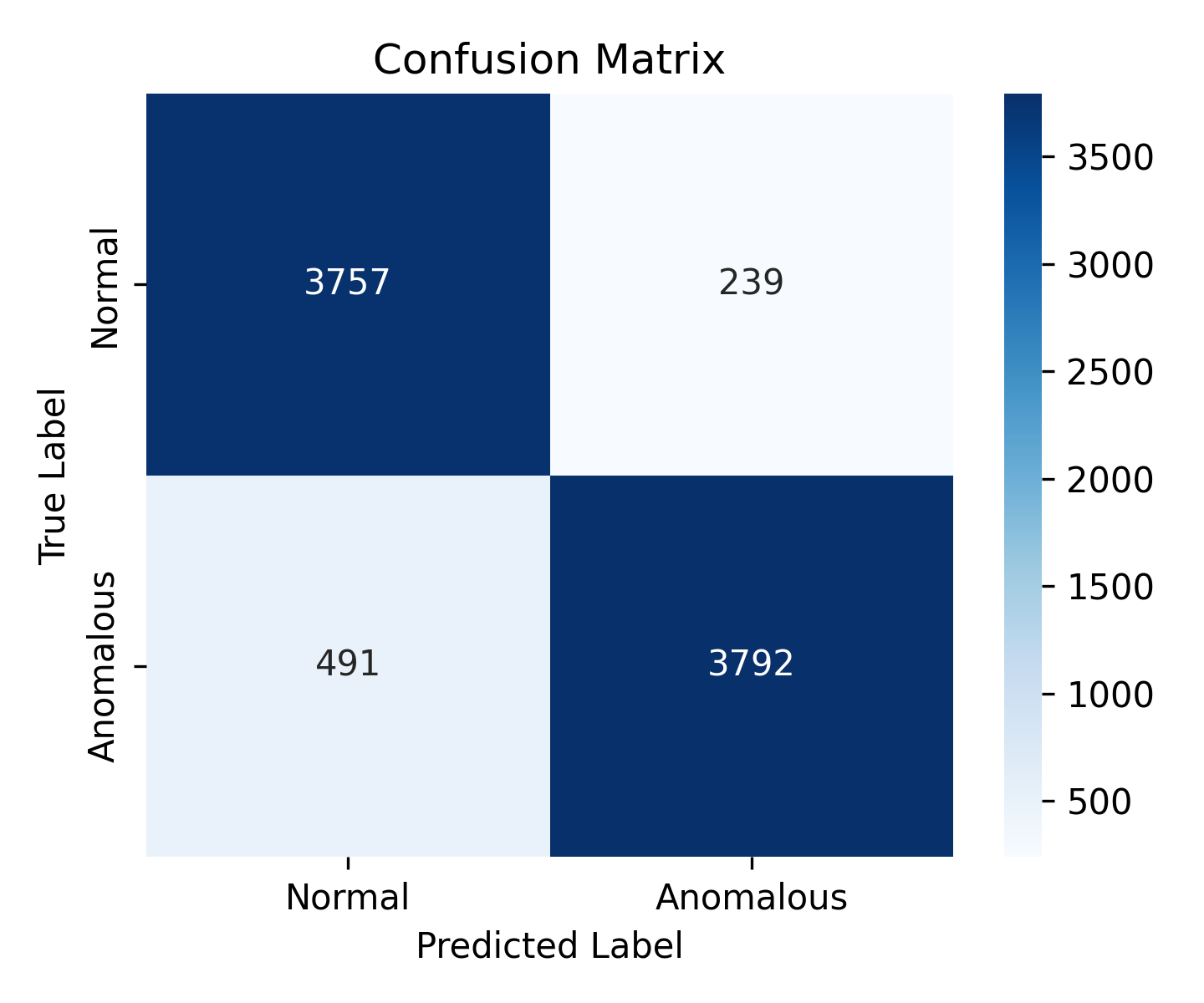}
    \caption{Confusion matrix showing predicted vs. actual labels for the binary classification of urban sounds.}
    \label{fig:confmatrix}
\end{figure}

\begin{figure}[htbp]
    \centering
    \includegraphics[width=0.9\linewidth]{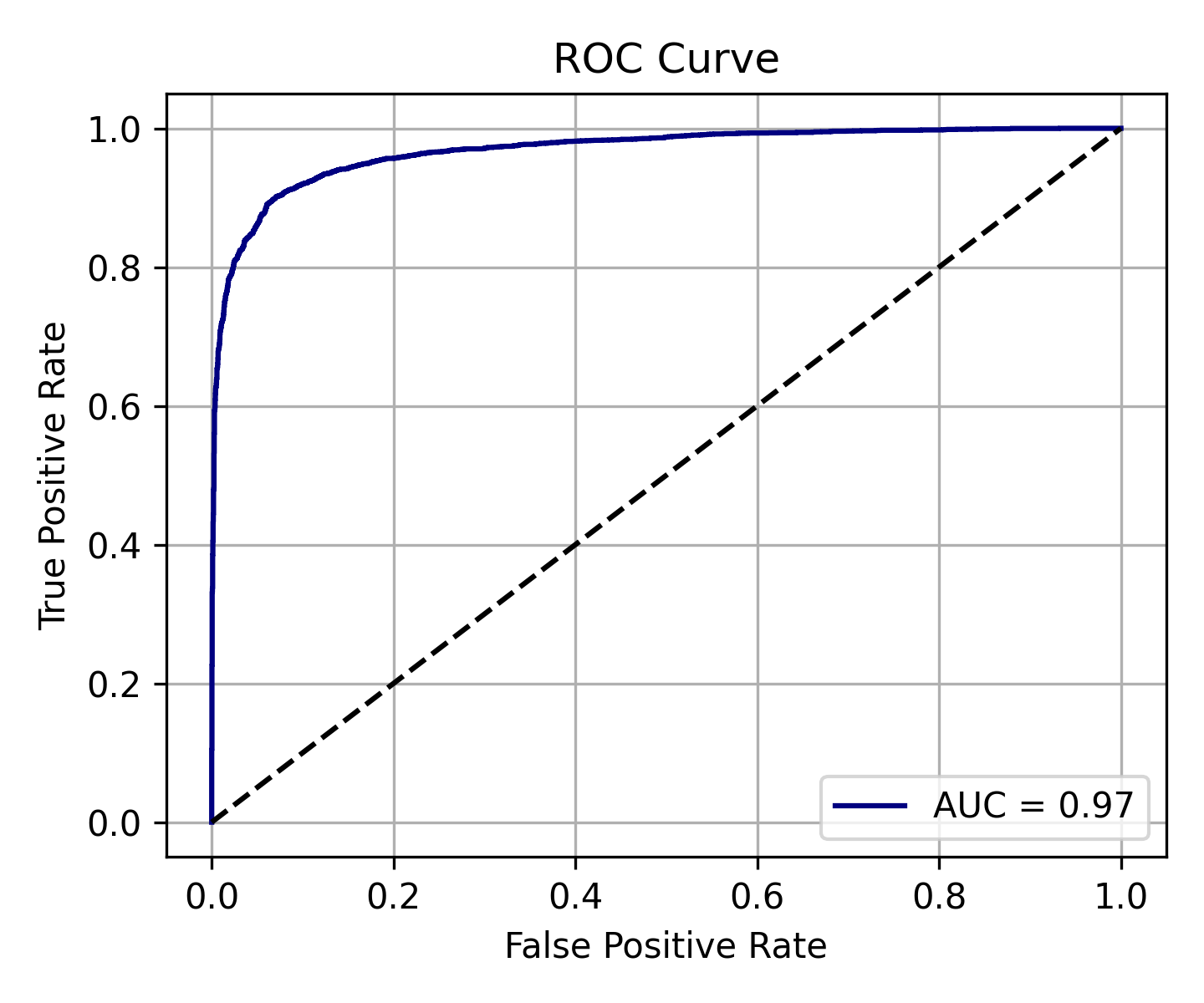}
    \caption{ROC curve showing the model’s ability to separate normal and anomalous audio events.}
    \label{fig:roc}
\end{figure}

\begin{figure}[htbp]
    \centering
    \includegraphics[width=0.9\linewidth]{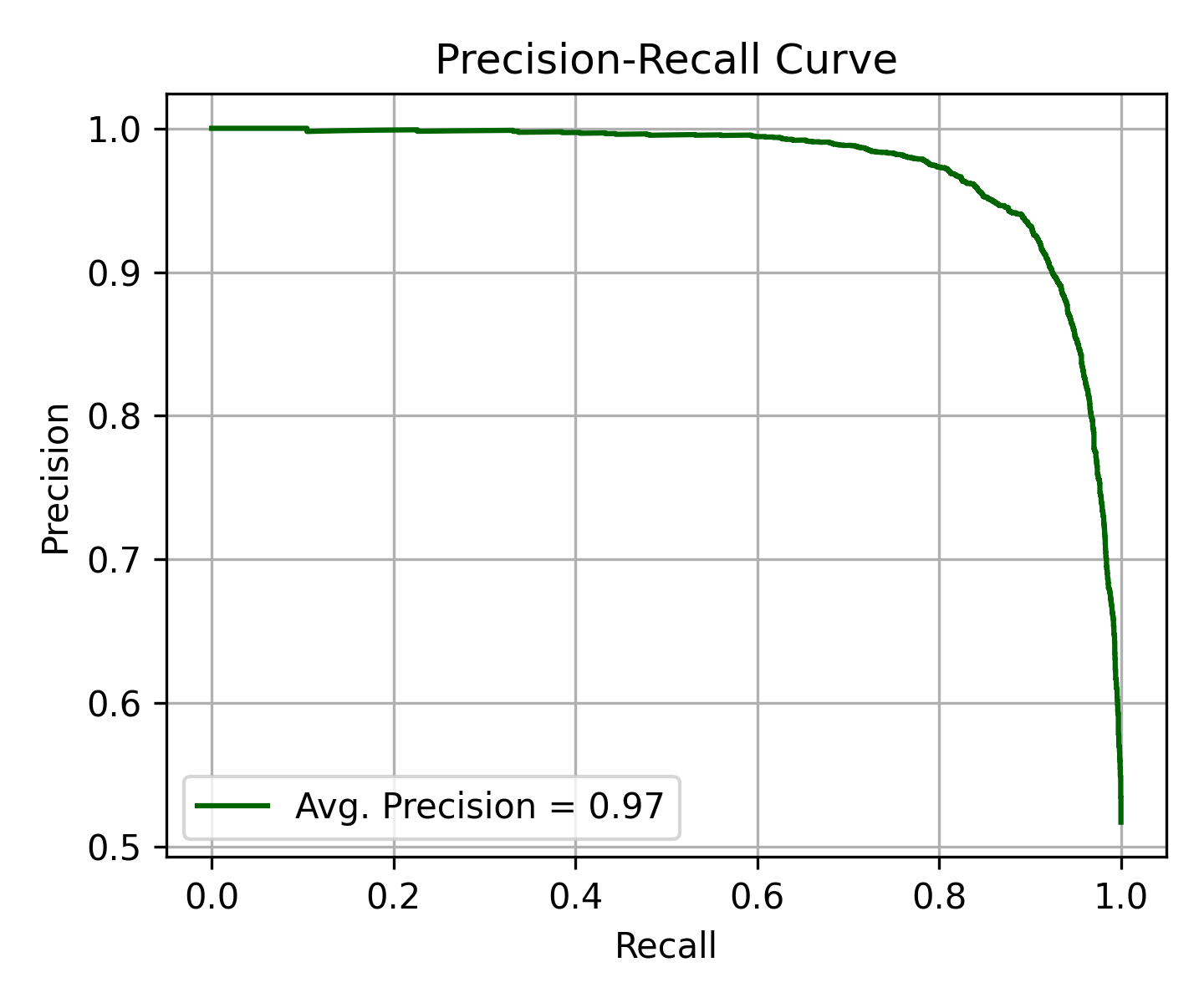}
    \caption{Precision-Recall curve highlighting the trade-off between detection sensitivity and alert specificity.}
    \label{fig:pr}
\end{figure}

\begin{figure}[htbp]
    \centering
    \includegraphics[width=0.9\linewidth]{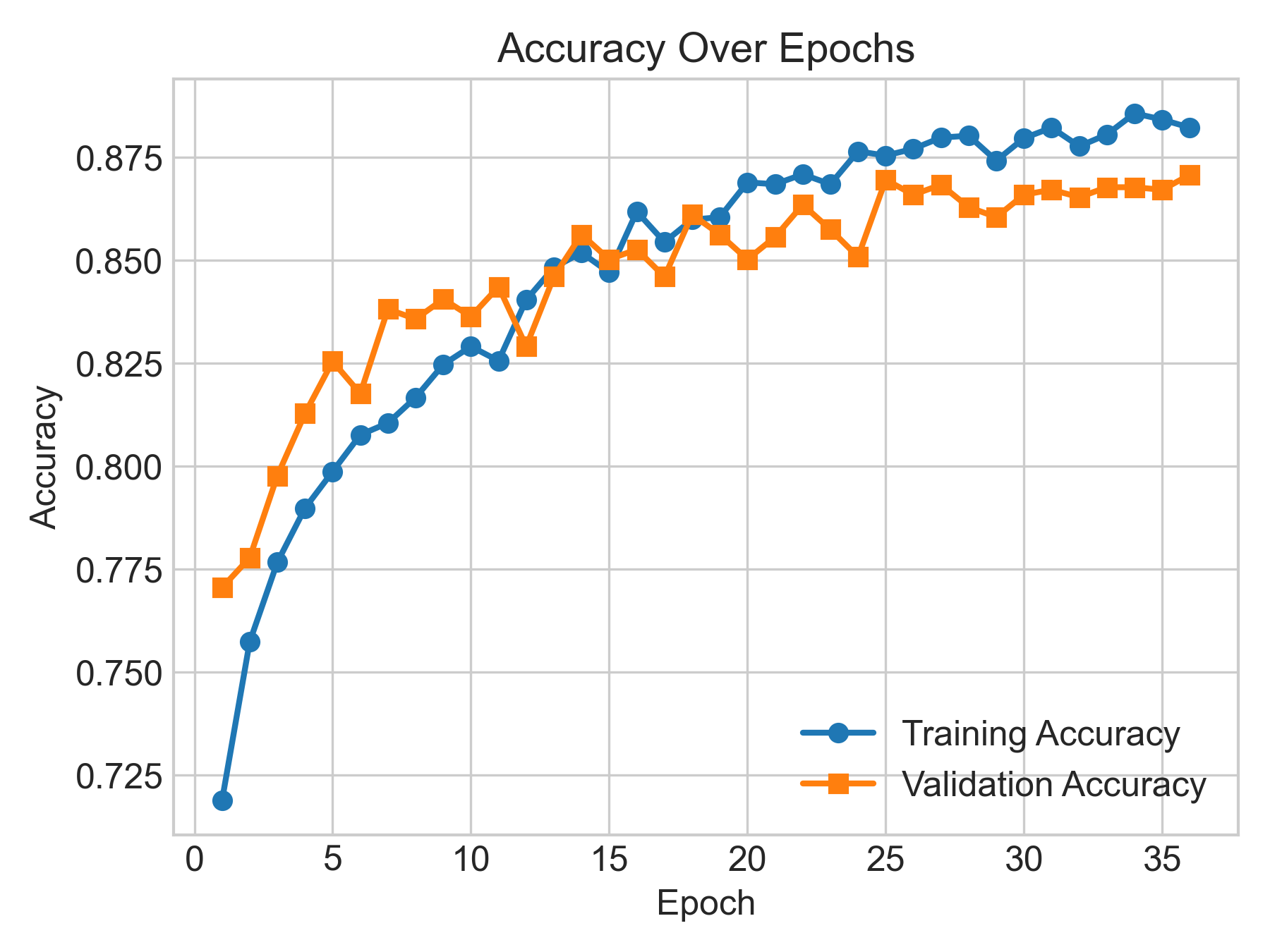}
    \caption{Training and validation accuracy across 35 epochs. Early stopping triggered after performance plateaued.}
    \label{fig:acc}
\end{figure}

\begin{figure}[htbp]
    \centering
    \includegraphics[width=0.9\linewidth]{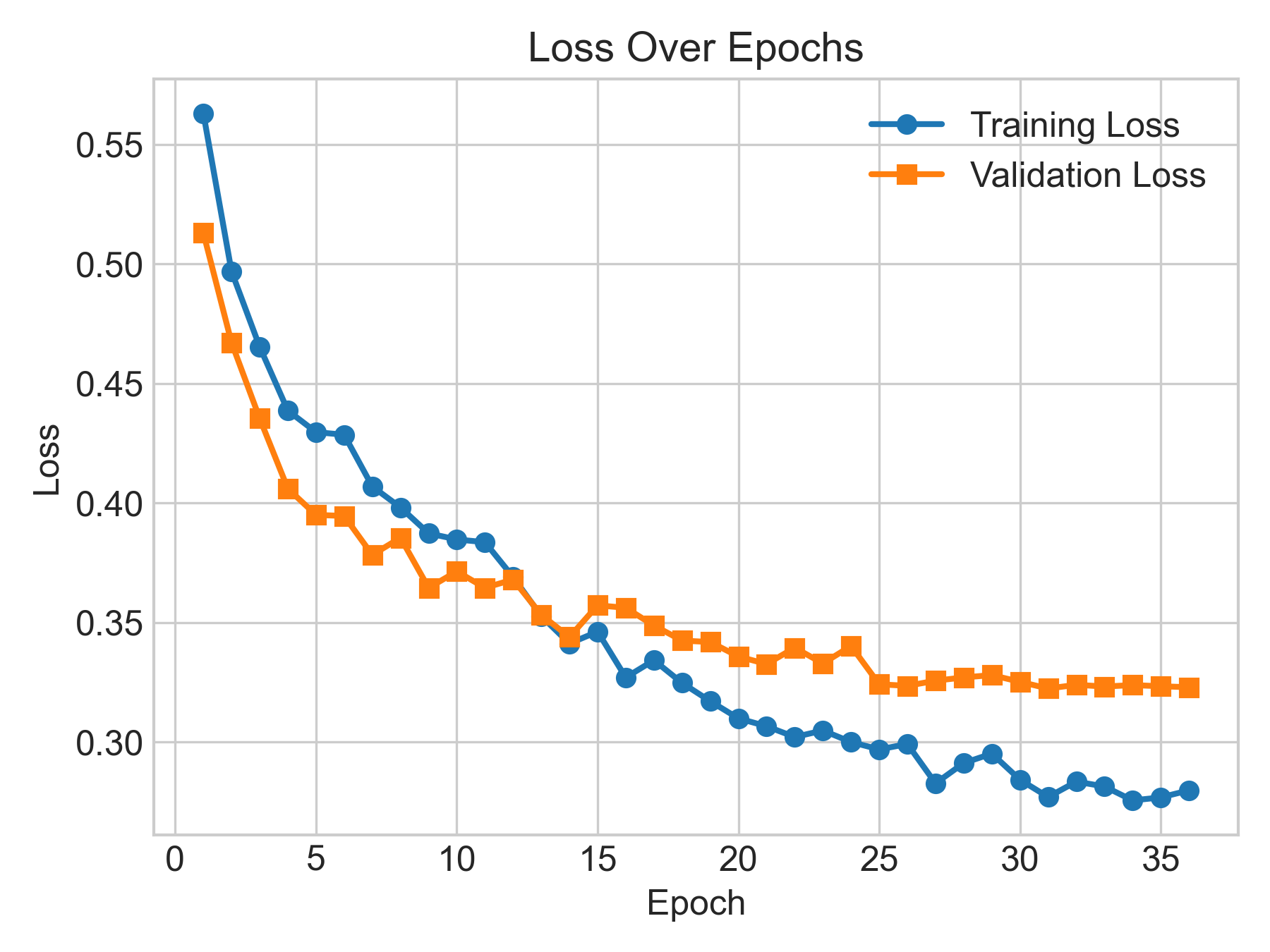}
    \caption{Training and validation loss across 35 epochs. The model exhibits stable learning and generalization.}
    \label{fig:loss}
\end{figure}

\noindent Detailed evaluation plots for the quantized model are shown in Figs.~\ref{fig:confmatrix} to \ref{fig:loss}. For the original (unquantized) model, we omit figures for brevity, as the patterns are similar but slightly stronger discriminative trends.  The confusion matrix (Fig.~\ref{fig:confmatrix}) confirms that the model effectively distinguishes between normal and anomalous events. Further evaluation using ROC and Precision-Recall curves (Figs.~\ref{fig:roc} and~\ref{fig:pr}) demonstrates high discriminative ability, with strong AUC and average precision scores. Training dynamics are visualized in Figs.~\ref{fig:acc} and~\ref{fig:loss}, showing stable convergence and no signs of overfitting. These results confirm that compact neural networks trained on MFCC features can perform accurate and reliable anomaly detection in audio streams. Despite their small footprint, such models generalize well across diverse urban environments. The confusion matrix and class-wise metrics indicate consistent performance across both normal and anomalous classes. This balance is especially valuable in urban safety and monitoring applications, where false alarms (low precision) and missed anomalies (low recall) both carry practical costs. Importantly, the model achieves this level of performance using a lightweight architecture suitable for microcontroller deployment. Compared to prior embedded approaches such as~\cite{rani2022noiseclassification} and~\cite{hammad2023unsupervised}, our system delivers competitive results while remaining fully compatible with TensorFlow Lite for Microcontrollers. The high AUC and average precision further underscore the model’s discriminative capability across decision thresholds. This flexibility could support threshold tuning or cost-sensitive optimization in future work, allowing adaptation to application-specific constraints. For example, maximizing recall may be prioritized in safety-critical environments. Misclassifications, when they occur, often involve overlapping spectral content between ambient and anomalous sounds (e.g., street music misclassified as sirens). Reducing such confusion may require advanced feature engineering, ensemble models, or hybrid detection-classification pipelines. Nonetheless, the system’s overall accuracy, efficiency, and generalizability highlight the feasibility of embedding sound-based intelligence directly on IoT nodes without reliance on cloud-based inference. While the UrbanSound8K dataset provides a practical benchmark for controlled evaluation, future work will include validation on real-world IoT sensor network data to assess robustness under diverse acoustic and environmental conditions.

\section{Conclusion and Future Work} \label{Conclusion}
This paper presented a complete TinyML pipeline for real-time acoustic anomaly detection in IoT sensor networks. Using MFCC-based features and a compact neural network, the system was trained and evaluated on the UrbanSound8K dataset, achieving 91\% accuracy and strong precision-recall balance across both normal and anomalous classes. The architecture was designed with embedded deployment in mind and remains fully compatible with TensorFlow Lite for Microcontrollers. Evaluation metrics and training curves confirmed that the model generalizes well, without signs of overfitting, and offers robust classification performance suitable for resource-constrained environments. The proposed approach demonstrates that accurate sound classification can be achieved on-device without reliance on cloud resources, enabling privacy-preserving and responsive monitoring in smart cities, buildings, and safety-critical infrastructure. Misclassifications, while limited, highlight the need for improved spectral discrimination and contextual awareness in certain edge cases. Future work will focus on deploying the trained model on actual microcontroller hardware and evaluating real-time performance under memory and inference constraints. In addition, we plan to explore further model compression techniques such as pruning and knowledge distillation to reduce computational overhead. We also plan to evaluate on-device memory usage, inference latency, and power consumption to provide concrete insights into practical feasibility for battery-powered IoT deployments. Expanding the system toward multi-class classification and continuous streaming inference is also a promising direction for supporting richer and more adaptive edge-based acoustic intelligence.

\balance
\bibliographystyle{IEEEtran}  
\bibliography{references}     

\end{document}